\documentclass[a4paper,twocolumn]{article}

\usepackage{jsaiac}

\usepackage{color}
\usepackage{subcaption}
\usepackage{microtype}
\usepackage{multirow}
\usepackage{graphicx}
\usepackage{booktabs}
\usepackage{url}
\usepackage{stfloats}
\usepackage{makecell}
\usepackage[english]{babel}
\usepackage[square,numbers]{natbib}
\usepackage{longtable} 
\usepackage{lipsum}
\title{A Better LLM Evaluator for Text Generation: \\The Impact of Prompt Output Sequencing and Optimization}

\address{\{kcchu,ypc\}@nlab.ci.i.u-tokyo.ac.jp}

\author{%
KuanChao Chu\textsuperscript{*}
\and
Yi-Pei Chen\textsuperscript{*}
\and
Hideki Nakayama
}

\affiliate{
Graduate School of Information Science and Technology, The University of Tokyo
}

\begin{abstract}
This research investigates prompt designs of evaluating generated texts using large language models (LLMs). While LLMs are increasingly used for scoring various inputs, creating effective prompts for open-ended text evaluation remains challenging due to model sensitivity and subjectivity in evaluation of text generation. Our study experimented with different prompt structures, altering the sequence of output instructions and including explanatory reasons. We found that the order of presenting reasons and scores significantly influences LLMs' scoring, with a different level of rule understanding in the prompt. An additional optimization may enhance scoring alignment if sufficient data is available. This insight is crucial for improving the accuracy and consistency of LLM-based evaluations.
\end{abstract}

\def\BibTeX{{\rm B\kern-.05em{\sc i\kern-.025em b}\kern-.08em%
 T\kern-.1667em\lower.7ex\hbox{E}\kern-.125emX}}
\def\JBibTeX{\leavevmode\lower .6ex\hbox{J}\kern-0.15em\BibTeX}
\def\LaTeXe{\LaTeX\kern.15em2$_{\textstyle\varepsilon}$}

\begin{document}
\maketitle

\section{Introduction}
\begingroup
\renewcommand\thefootnote{}
\footnotetext{Equal contribution}
\endgroup

Using large language models (LLMs)~\cite{GPT4,llama2} as evaluators to assign scores to the given inputs have become prevalent. 
\citet{AlphaCode2} outputs a score between 0 and 1 to estimate the correctness of generated code, thereby automatically ranking its quality. Similarly, \citet{generative_agent} assign poignancy score to the generated text for the retrieval task. 
Other research explores using LLMs to assess generated texts, finding the LLM scores correlates higher with human evaluators than existing automatic metrics \cite{gao2023human,shen2023large,liu-etal-2023-g,luo2023chatgpt}. 

However, designing evaluation prompt for LLMs is not a trivial task, especially for dialogue evaluation. 
Different models exhibit varied sensitivity to the nuances of input prompts. Even slight linguistic variations can lead to significant fluctuations in task performance \cite{leidinger2023linguistic}. 
Moreover, the inherent subjectivity in dialogue evaluation adds on the difficulty and versatility in LLMs' evaluation results. 
While prompt optimization techniques \cite{chen2023instructzero,LLMoptim,zhang2023tempera,prasad2023grips} have been developed to assist in designing more effective prompts, these methods require paired input-output samples for objective value calculation.  
Unfortunately, the lack of available dialogue-score pairing data hampers the application of prompt optimization in dialogue evaluation. 

In this study, we aim to investigate the influence of prompt design on dialogue evaluation, specifically focusing on how the output instructions affects the resulting scores. 
We have developed multiple prompt variations to assess the quality of a series of dialogues.  
These variations involve altering the sequence order of the outputs and examining whether including explanatory reasons along with the scores impacts the evaluation. 
Our analysis compares the influence of different prompts on the scoring outcomes across various versions of GPT models. 

\begin{figure}[t]
\centering

\begin{subfigure}{0.5\columnwidth}
  \includegraphics[width=\linewidth]{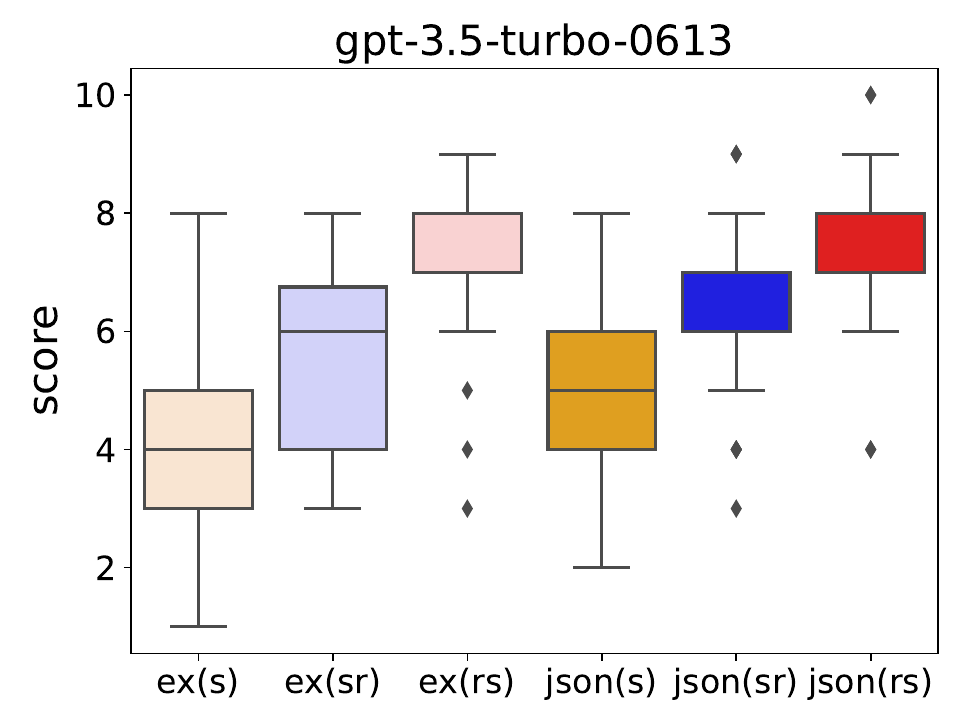}
\end{subfigure}%
\begin{subfigure}{0.5\columnwidth}
  \centering
  \includegraphics[width=\linewidth]{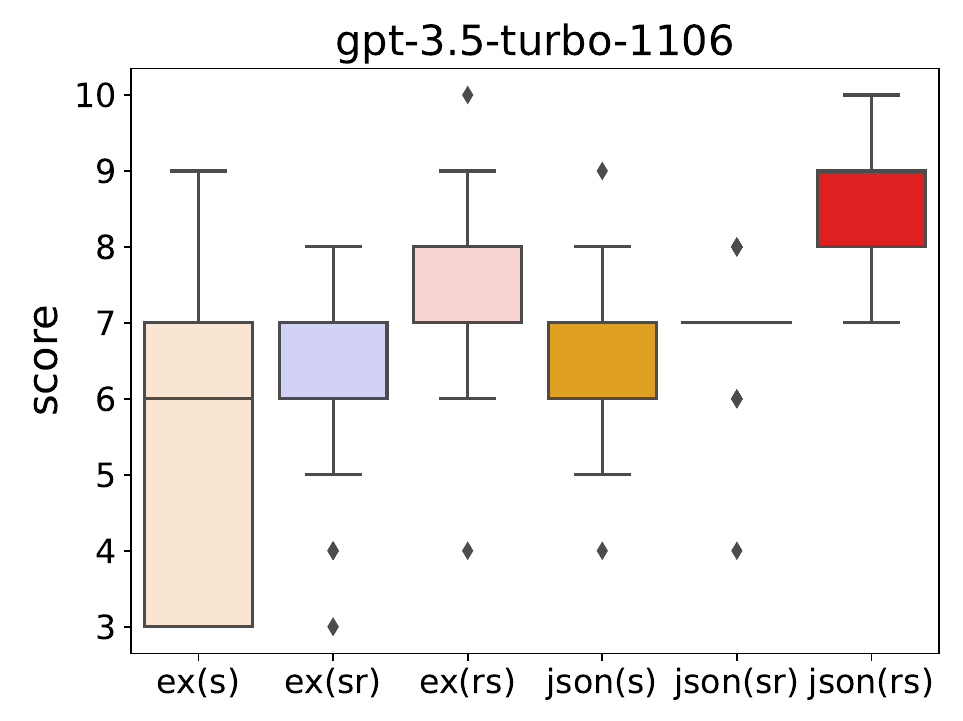}
\end{subfigure}

\begin{subfigure}{0.5\columnwidth}
  \centering
  \includegraphics[width=\linewidth]{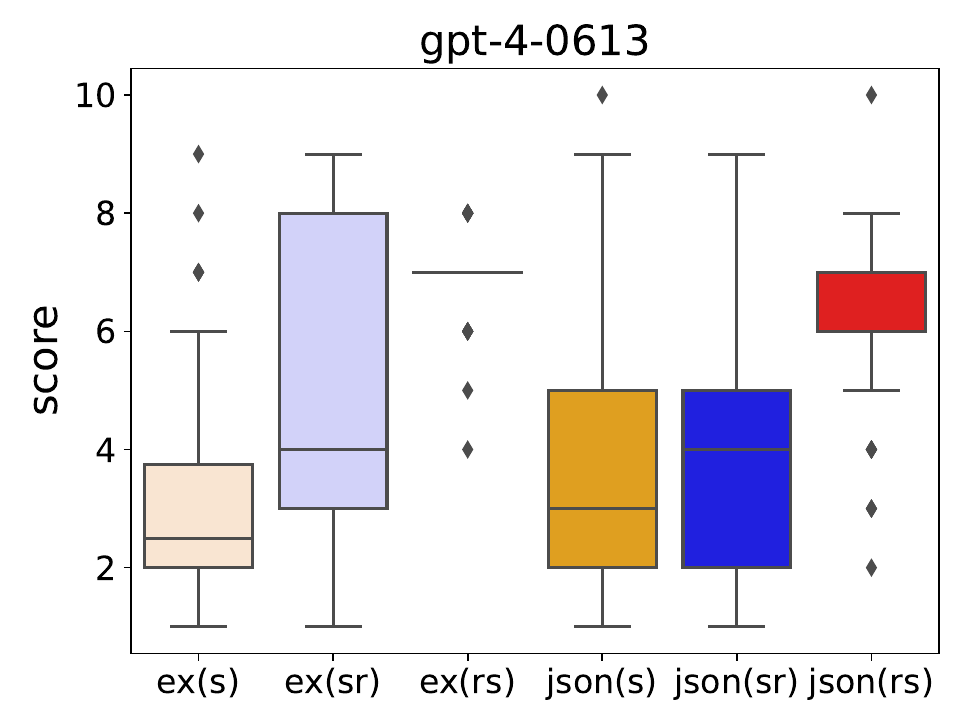}
\end{subfigure}%
\begin{subfigure}{0.5\columnwidth}
  \centering
  \includegraphics[width=\linewidth]{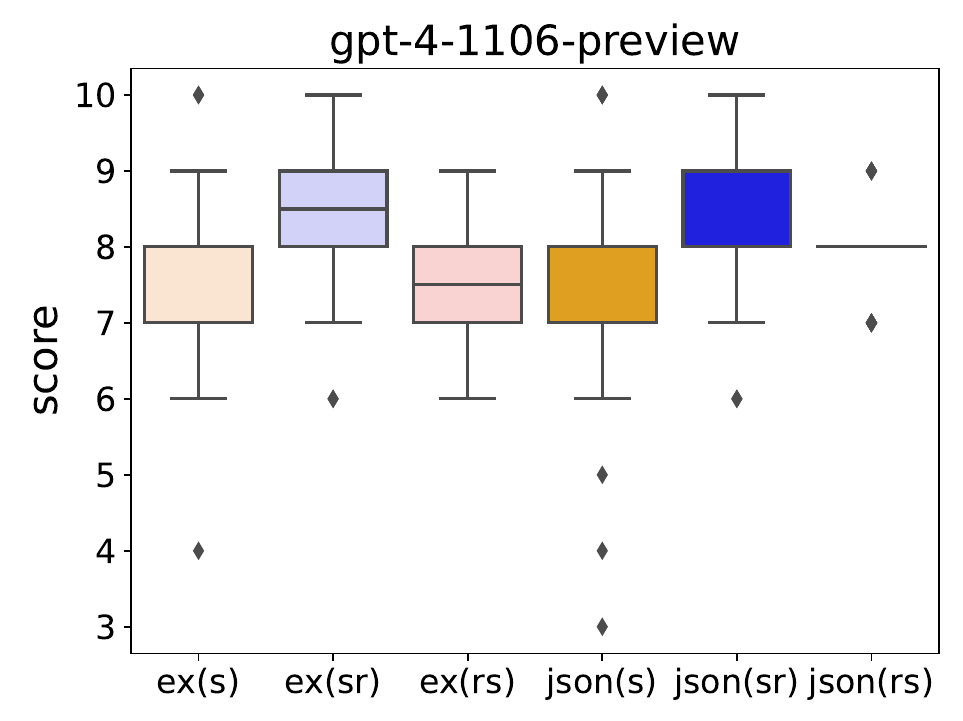}
\end{subfigure}

\caption{Score distribution across 50 trials for each model and output instruction configuration for a dialogue set.}
\label{fig:box_case22}
\end{figure}

We observed that the different order of output instructions can result in different scoring distributions by certain LLMs, even when the corresponding output reasons are similar.  
Considering the sequential generation nature of auto-regressive models, placing the score after the reasons allows it to reference both the reasons and the input prompt, a dynamic not possible when this order is reversed.  
The finding suggests that a ``reason-first'' output instruction might lead to a more comprehensive understanding and adherence to the specific requirements of the task. 

Finally, we further investigate the effect of prompt optimization on the ``reason-first'' output instruction. This is to observe the most effective form of instruction in a hypothetical setting where paired data is available.

\begin{figure*}[t]
\centering
\includegraphics[trim={0cm 0.75cm 0cm 2.7cm},clip,width=\linewidth]{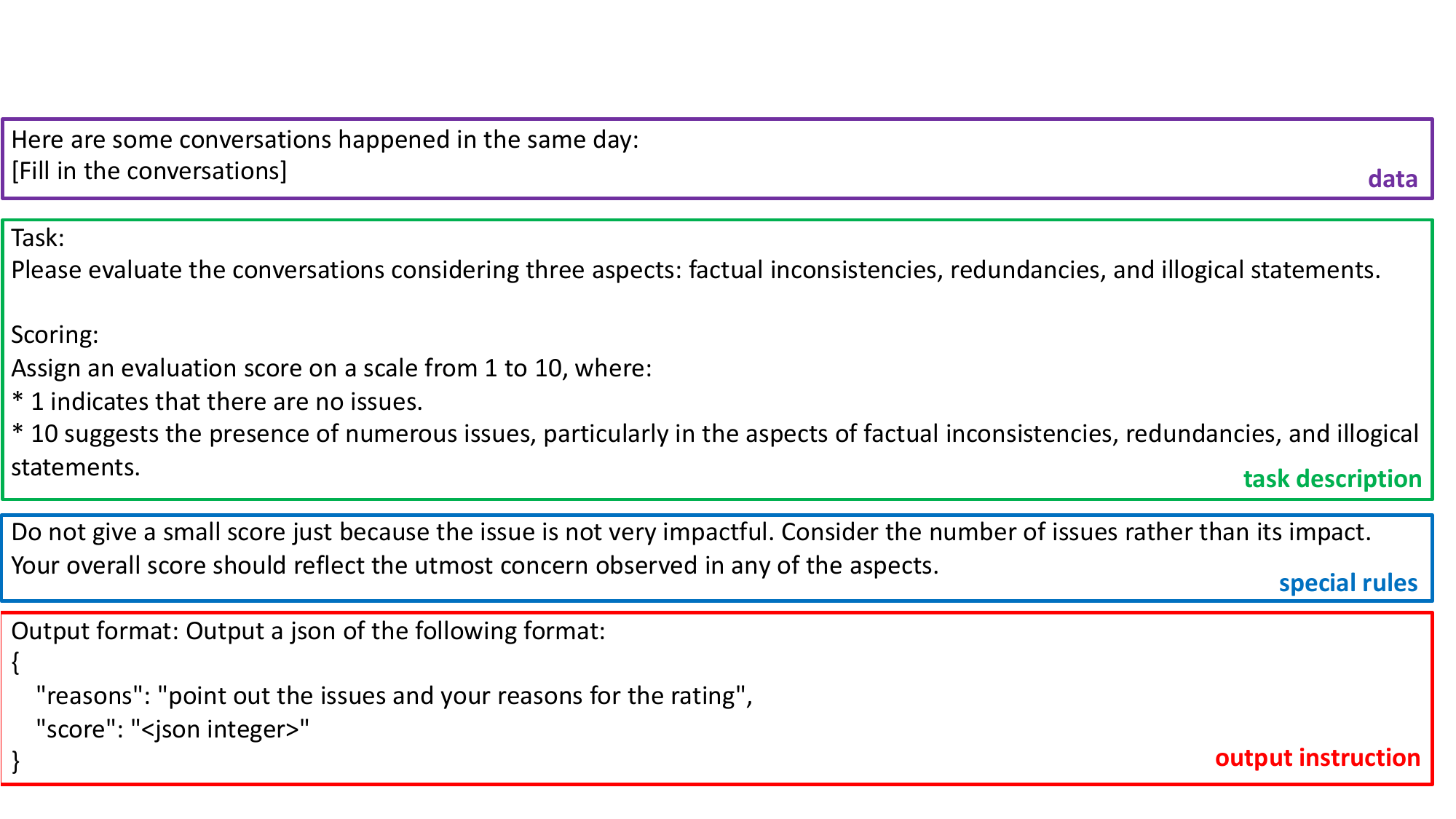}
\caption{The form of prompt for LLM scorer in conversation evaluation. The special rules section is simplified for better readability.}
\label{fig:method}
\end{figure*}

\section{Output Instruction in Dialogue Evaluation}
\label{sec:order}

\begin{table}[h]
\centering
\small
\resizebox{1.0\linewidth}{!}{
\begin{tabular}{lp{0.78\linewidth}}
\toprule
Config & Output Instruction in the Prompt \\
\midrule
ex (s) & \texttt{Example JSON output:}\\
      & \texttt{\{"score": 5\}}\\
\midrule          
ex (sr)   & \texttt{Example JSON output:}\\
          & \texttt{\{"score": 5, "reasons": "$<$your reasons for the rating$>$"\}}\\
\midrule 
ex (rs)   & swap the order of ``score'' and ``reasons'' in ex (sr)\\
\midrule
json (s) & \texttt{Output a json of the following format:} \\
         & \texttt{\{"score": "$<$integer$>$"\}}\\
\midrule
json (sr) & \texttt{Output a json of the following format:}\\
          & \texttt{\{"score": "$<$integer$>$", "reasons": "point out the issues and your reasons for the rating"\}} \\
\midrule
json (rs) & swap the order of ``score'' and ``reasons'' in json (sr)\\
\bottomrule
\end{tabular}
}
\caption{The variations of output instruction.}

\label{tab:out_instruct}
\end{table}
\paragraph{Approach}
\hspace*{0.5em}
In this section, the task assigned to the LLM is to rate a given set of dialogues on a scale from 1 to 10, where 1 indicates no issues in the set of dialogues, and 10 signifies severe problems. 
Additionally, if specified in the prompt, the LLM is required to provide a rationale for the rating. 
The dialogues are presented in chronological order, and the output score is determined based on a comprehensive evaluation of the entire set, focusing on key aspects such as factual accuracy, repetitiveness, and coherence.

Along with the task description, we have integrated five customized rules into the prompt, derived from observations in previous experiments without these rules. The special rules include instructions for the LLM to prioritize the number of issues over their impact and to assign more weight to aspects exhibiting significant issues, rather than averaging out the score across all aspects.

The final evaluation prompt is organized as follows: a set of dialogues, task description, special rules, and output instruction (see Table~\ref{tab:out_instruct}).
For each set of dialogues, we conducted $N$ trials for each of the six configurations (config), varying the output instruction. This experiment was then replicated across $M$ different models.

\begin{figure}[h]
\centering

\begin{subfigure}{0.5\columnwidth}
  \includegraphics[width=\linewidth]{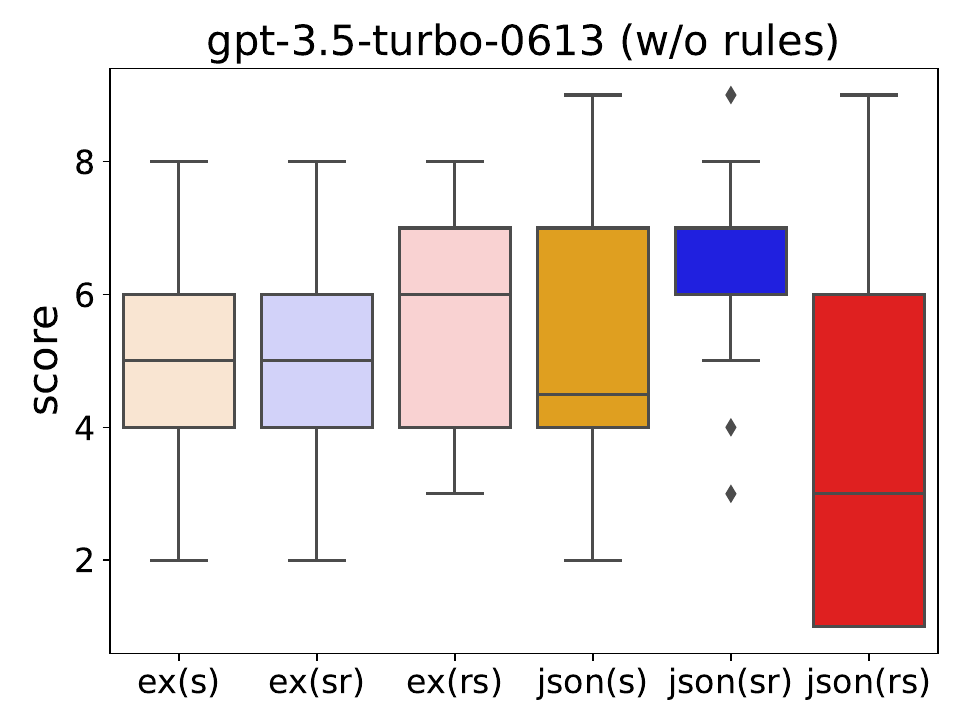}
\end{subfigure}%
\begin{subfigure}{0.5\columnwidth}
  \centering
  \includegraphics[width=\linewidth]{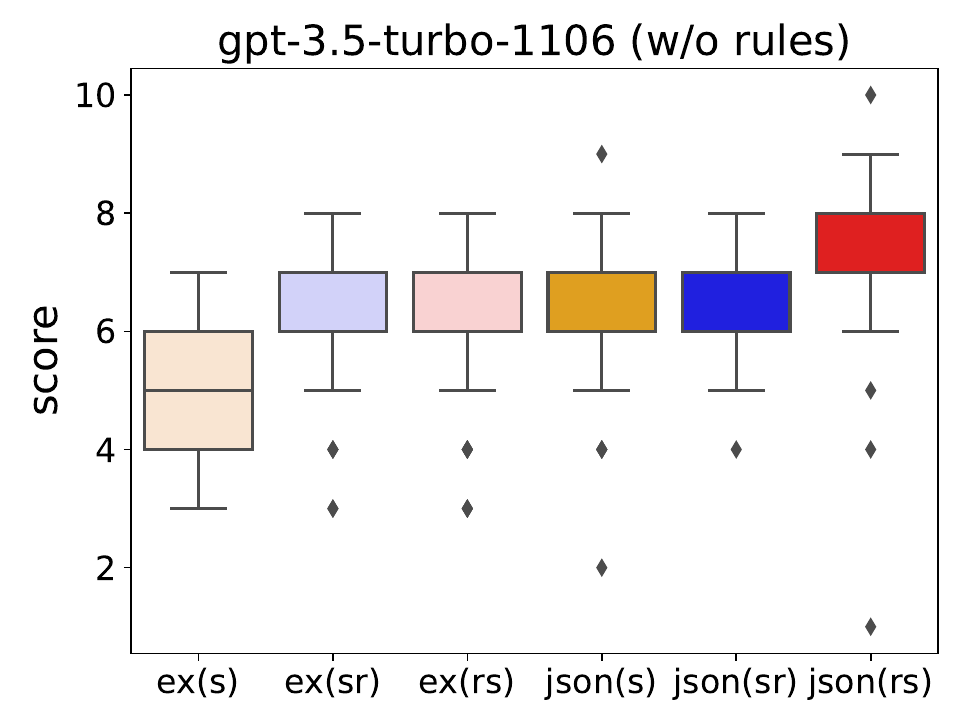}
\end{subfigure}

\begin{subfigure}{0.5\columnwidth}
  \centering
  \includegraphics[width=\linewidth]{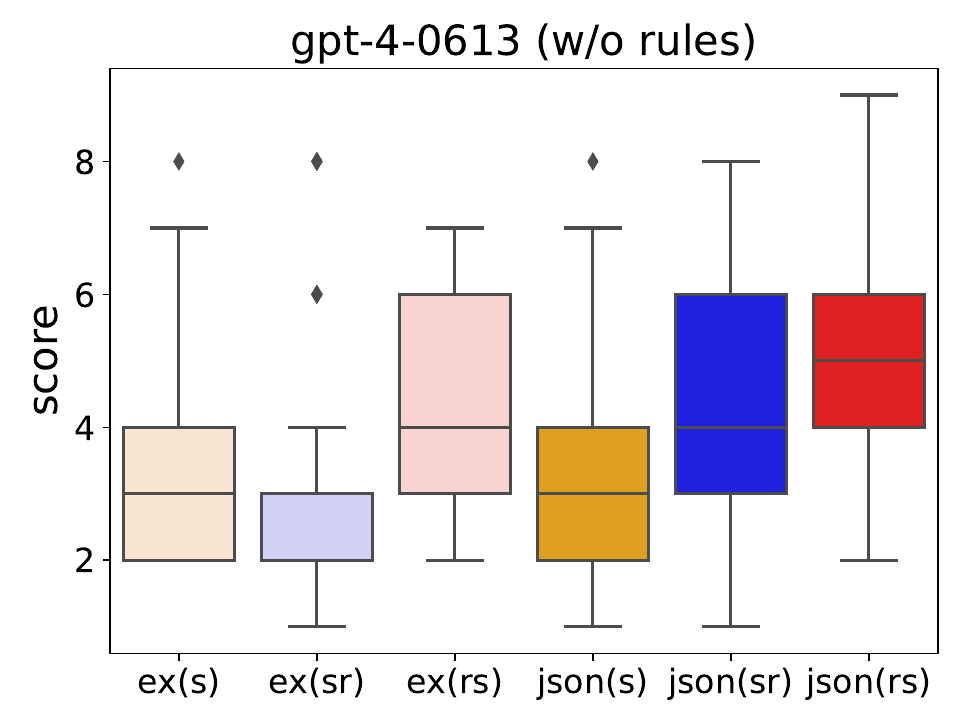}
\end{subfigure}%
\begin{subfigure}{0.5\columnwidth}
  \centering
  \includegraphics[width=\linewidth]{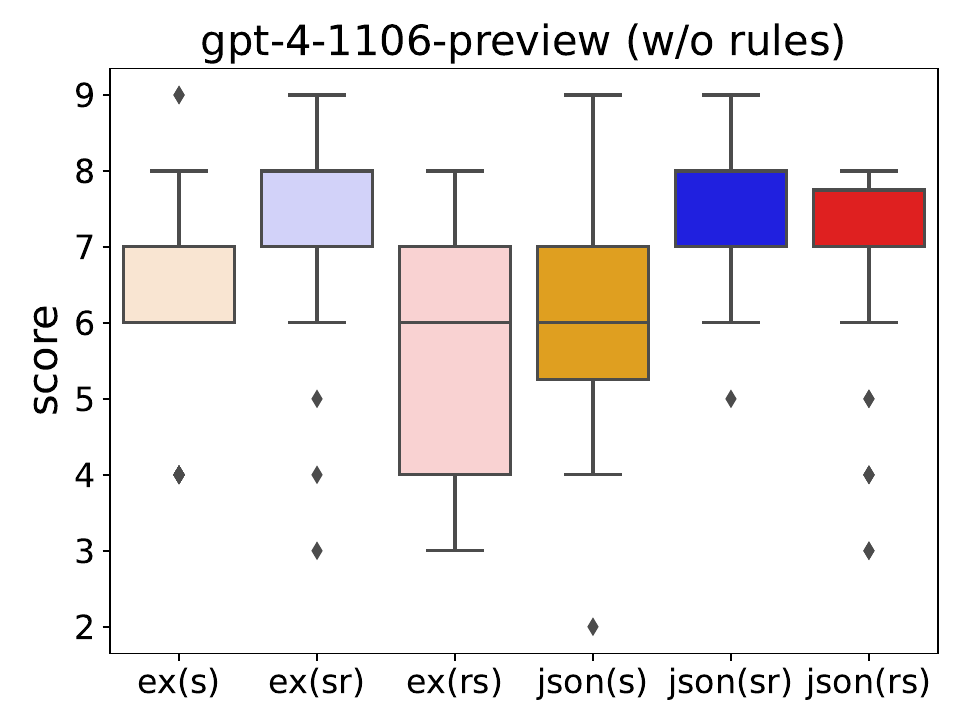}
\end{subfigure}

\caption{
Score distribution across 50 trials for each model and output instruction configuration for a dialogue set, with the `special rules' omitted from the prompt.
}
\label{fig:box_case22_norule}
\end{figure}
\paragraph{Data}
\hspace*{0.5em}
To assess the capability of LLMs in identifying issues within dialogues, we collected LLM-generated dialogues from \citet{generative_agent} and manually grouped them into 25 sets. Each set contains four to six dialogues and exhibits one or more problems, such as repetition or contradictions between dialogues. 

\paragraph{Model}
\hspace*{0.5em}
We selected four recent LLMs to serve as scorers: gpt-3.5-turbo-0613, gpt-3.5-turbo-1106, gpt-4-0613, and gpt-4-1106-preview (gpt-4-1106). 
Note that our aim is to analyze the evaluation scores across various models when altering output instructions, and not to compare them with human judgements for this task. 

\subsection{Result and Analysis}

\begin{table}[h]
\centering
\small
\tabcolsep 4pt
\resizebox{1.0\linewidth}{!}{
\begin{tabular}{lcccc}
\toprule
\multirow{2}{*}{Config} &\multicolumn{2}{c}{GPT-3.5-turbo} &\multicolumn{2}{c}{GPT-4}\\
\cmidrule(lr){2-3}\cmidrule(lr){4-5}
  & -0613 & -1106  &-0613 & -1106 \\
\midrule  
ex (s) & 3.68 $\pm$1.17 & 4.51 $\pm$1.19& 3.36 $\pm$1.07 & \textbf{8.18} $\pm$1.05\\

ex (sr) & 4.20 $\pm$1.19 & 5.49 $\pm$1.22 & 3.39 $\pm$1.13 & 7.55 $\pm$1.12\\

ex (rs) & \textbf{6.09} $\pm$1.23 & \textbf{7.66} $\pm$0.81 & \textbf{5.58} $\pm$1.19 & 7.39 $\pm$0.90\\
\midrule
json (s) & 4.03 $\pm$1.16 & 6.18 $\pm$1.09& 3.13 $\pm$1.10 & 6.74 $\pm$1.24\\

json (sr) & 4.66 $\pm$1.15 & 6.76 $\pm$0.94 & 3.26 $\pm$1.11& \textbf{7.69} $\pm$1.06\\

json (rs) & \textbf{5.78} $\pm$1.42 & \textbf{7.99} $\pm$0.94 & \textbf{5.34} $\pm$1.22 & 7.54 $\pm$0.95\\

\bottomrule
\end{tabular}
}
\caption{Mean scores and std for 25 dialogue sets, evaluated across different models and output instruction configurations.}

\label{tab:main}
\end{table}

\paragraph{The Importance of Output Instruction Order}
\hspace*{0.5em}

Table~\ref{tab:main} presents the mean scores and standard deviations (std) of 10 trials for all 25 dialogue sets across all configs and models. In both \textit{ex~($\cdot$)} and \textit{json~($\cdot$)} formats, the mean scores for the \textit{rs} settings (output reasons before the score) are generally higher than their \textit{sr} (output score before reasons) counterparts.~\footnote{The exception is observed with the gpt-4-1106 model.}
For instance, in the \textit{json~(rs)} config using gpt-4-0613, the mean score is 5.34, while it drops to 3.26 in \textit{json~(sr)}, despite providing similar reasons. 
We conjecture that in the \textit{rs} setting, the autoregressive nature of the model allows the score to be influenced by the previously outputted reasons.

\paragraph{Different Levels of Rule Understanding}
\hspace*{0.5em}
In a focused study on a single set with additional 40 trials, as depicted in Fig.~\ref{fig:box_case22}, we observed a trend consistent with the findings presented in Table~\ref{tab:main}. 
However, as shown in Fig.\ref{fig:box_case22_norule}, when we removed the `special rules' from the prompt, we found that most scores were lower and the distinctions between different settings became less pronounced.
This highlights the models' sensitivity to the changes of the prompt.

\begin{table*}[tp]
\centering
\small
\tabcolsep 4pt
\resizebox{1.0\linewidth}{!}{
\begin{tabular}{lclcc}
\toprule
Method & Iteration & Output Instruction & \multicolumn{2}{c}{score on $\mathcal{S}$ ($\uparrow$)} \\
\cmidrule(lr){1-3} \cmidrule(lr){4-5}
init(json(rs))& 0 & Output a json of the following format: \{"reasons": "point out your reasons for the rating on coherence", "score": "the rating"\}  & -1.271 & 51.9 \\

\midrule
+GRIPS & 1 & Output a json of \hspace{0.05cm} following format: \{"reasons": "point out your reasons for \hspace{0.05cm} rating on coherence", "score": "the rating"\} & -1.267 & \\
      & 4 & Output json of following format: \{"reasons": "point out your reasons for rating on coherence", a "score": "of rating"\} & -1.234 & \\
      & 7 & Output json of following format "\{"reasons" ""point out your reasons for rating on coherence", "score" ""of rating"\} & \textbf{-1.100} & \\
\midrule
+OPRO & 25 & \makecell[l]{Assess the provided summary's coherence in relation to the source text, grading it on a scale of 1 to 5. \\Coherence refers to the seamless connection of ideas and the logical structuring of information from the source into a unified summary. \\The summary should not just be a cluster of related information, but a well-knit narrative reflecting the source's key points. \\Ensure you provide concise reasoning for your rating, discussing the summary's strengths or weaknesses in its coherence. \\Consistently refer back to these guidelines during your assessment.} & & 46.0\\
\cmidrule(lr){2-3}
&43 & \makecell[l]{Evaluate the summary provided, grading its coherence with the original news article on a scale of 1-5.\\ Consider how the summary reflects the logical progression, structure, and key details from the source text.\\ A superior summary should interpret related details into a clear, comprehensive story, not merely gathering related facts. \\For an effective assessment, underpin your score with explicit examples from the text.\\ Concentrate on the flow and organization of the summary throughout your evaluation.} & & \textbf{62.2} \\

\bottomrule
\end{tabular}
}
\caption{Examples of output instructions with optimization methods GRIPS and OPRO. We pick the one with highest scores on $\mathcal{S}$ as the optimized instructions.}

\label{tab:optimprompt}
\end{table*}

\section{Prompt optimization}

\paragraph{Approach}
\hspace*{0.5em}
We assume there are sufficient paired input and output data samples for the scoring task. Hence, in this section, we examine how output instructions can be enhanced through additional prompt optimization. We apply two recent optimization methods to the existing prompt: GRIPS \cite{prasad2023grips} and OPRO \cite{LLMoptim}. For a given score set $\mathcal{S}$ (i.e., input and output data pairs), GRIPS iteratively searches for a chain of editing commands to maximize the overall score on $\mathcal{S}$. Meanwhile, OPRO directly employs a LLM as an optimizer to generate new prompts, with previous score trajectories included as part of the reference material for the optimizer. Specifically, we utilize the \textit{json~(rs)} style for the output instruction in the initial prompt, \textbf{optimizing only the output instruction section while keeping the other sections unchanged.}

\paragraph{Data and Implementation Details}
\hspace*{0.5em}
For this experiment, we utilize SummEval \cite{fabbri2021summeval}, which comprises 1,600 text summarization samples derived from 100 unique document pieces and 16 summarization models. Each sample includes human expert ratings ranging from 1 to 5 across various aspects. We focus on the "coherence" aspect for the expected output. To ensure rating diversity in $\mathcal{S}$, we uniformly sample 10\% of the documents after ranking them by their average coherence score to create the score set for GRIPS, and use a halved subset for OPRO. The remaining 90\% of the data is reserved for testing.

For GRIPS, we configure the editing operations at the word level and conduct 10 search iterations, with 5 candidates in each iteration. For OPRO, we employ gpt-4-0613 as the optimizer, running 50 iterations with 2 generations each. We include two randomly selected data exemplars from $\mathcal{S}$ when generating the output instructions.

We calculate the mean absolute error (MAE) between the human ratings and LLM ratings to evaluate the overall score for $\mathcal{S}$. Furthermore, an additional entropy term is introduced for the binned output predictions to enhance output diversity, as suggested in the original paper, with a coefficient of $0.25$. Consistent with default settings, we linearly rescaled the score to a 0 to 100 rating scale for OPRO, and inverted the sign for GRIPS. All the methods utilize gpt-3.5-turbo-0613 for generating LLM scorer outputs. In the score set, the temperature was set to $0.3$, and a rating is averaged over 10 trials. In the test set, we set the temperature to $1.0$, with the number of trials increased to 20.

\subsection{Result and Analysis}
\paragraph{The Changes in Output Instructions and Generalization on Test Data} 
\hspace*{0.5em}
In the prompt examples shown in Table~\ref{tab:optimprompt}, OPRO demonstrates a preference for developing a diverse and detailed output instruction section. Some of the content either repeats or adds new details to the grading policy and may explicitly request the provision of reasons. The changes made using GRIPS, however, are comparatively limited due to the smaller search space. For example, the candidate with the best score on $\mathcal{S}$ underwent only five editing commands: ``del-del-swap-swap-del.''

Table~\ref{tab:optim} lists the performance of the LLM scorer on the test set. Surprisingly, the output instruction modified with GRIPS yield improved scores, whereas OPRO shows the opposite. However, the number of iterations set for OPRO is fewer than what was used in the original paper, due to cost constraints. Therefore, its performance might be far from saturation. Following \cite{chiang-lee-2023-closer}, we apply William's test to Pearson's \textit{r}, and we found that the improvement with GRIPS over the initial instruction is statistically significant.

\begin{table}[h]
\centering
\small
\tabcolsep 4pt
\resizebox{1.0\linewidth}{!}{
\begin{tabular}{lccc}
\toprule
 & MAE($\downarrow$) & Pearson's r($\uparrow$) & Kendall's $\tau$ ($\uparrow$)\\
\midrule
init(json(rs)) & 0.739 & 0.599 & 0.439 \\
+GRIPS & \textbf{0.696} & \textbf{0.614} & \textbf{0.449} \\
+OPRO & 0.780 & 0.466 & 0.328 \\

\bottomrule
\end{tabular}
}
\caption{The MAE and dataset-level correlation coefficients between LLM scorer and human experts ratings on the test set, using inital and post-optimization output instructions.}

\label{tab:optim}
\end{table}

\paragraph{The Score Difference for Swapped Orders Exists}
\hspace*{0.5em}
We also compare the \textit{json~(rs)} and \textit{json~(sr)} types of initial instructions. The averaged score on the test set is 3.656 for \textit{json~(rs)} and 3.270 for \textit{json~(sr)}. This reflects a similar distribution difference to what we observed in the previous section. 

\paragraph{Challenges} 
\hspace*{0.5em}
First, although we intentionally divide a score set to study the effect of optimization, collecting related samples for a specific scoring task can be challenging. Second, these optimizations are applied globally to all data samples, while the optimized output instructions might vary for individual cases. Finally, the optimized instructions could break the structured output format and thus requires careful post-generation processing.

\section{Conclusion}
Our study highlights the scoring sensitivity of LLMs to the order of output instructions, which can be amplified by task-specific rules. 
Users should verify that their intentions align with the outputs using crafted prompts. 
Additionally, prompt optimization has proven beneficial in aligning LLM outputs with ground-truth scores, emphasizing the need for sufficient data samples for arbitrary scoring tasks. These findings offer insights for designing prompts in subjective tasks like dialogue evaluation.

\bibliography{ref}

\newpage
\onecolumn

\appendix
\section*{Appendix}
\section{Full Special Rules}

Table~\ref{tab:fullspecialrules} presents the full prompt of the special rules section in Fig.~\ref{fig:method}, which was trimmed for better readability.

\begin{table*}[thp]
    \centering
    \tabcolsep 4pt
    \resizebox{0.9\linewidth}{!}{
    \begin{tabular}{p{\textwidth}}
        \toprule
        "Unfinished conversation" need not be viewed as an issue.\\
        Do note that the scoring should not be an average of the performance in each of the three aspects.\\ 
        Even if the conversations are generally coherent but contain a issue in one aspect, the score should lean towards the higher end of the scale.\\ 
        Do not give a small score just because the issue is not very impactful. Consider the number of issues rather than its impact.\\
        Your overall score should reflect the utmost concern observed in any of the aspects.\\
        \bottomrule
    \end{tabular}
    }
    \caption{Special rules for conversation evaluation.}
    \label{tab:fullspecialrules}
\end{table*}

\section{Example Dialogue Set}

Table~\ref{tab:dialoguesample} lists an example dialogue set we have studied in Section~\ref{sec:order} The results shown in Fig.~\ref{fig:box_case22} and Fig.~\ref{fig:box_case22_norule} are produced from this set. The original data can be found in \cite{generative_agent}. \footnote{\url{https://reverie.herokuapp.com/arXiv_Demo/}}

\begin{longtable}{p{0.86\linewidth}}
        \toprule
        Time: 07:08\\
        Mei Lin: ``Good morning, John. How did you sleep?''\\
        John Lin: ``Not bad, although I had a lot on my mind. I'm curious about who will be running for the local mayor election next month.''\\
        Mei Lin: ``Oh, that's interesting. Have you asked anyone about it?''\\
        John Lin: ``Yes, I've been asking everyone I meet. I'm also taking online classes to stay up to date on new medications and treatments.''\\
        Mei Lin: ``That's great. By the way, have you noticed anything strange with Eddy lately? He seems a bit rebellious.''\\
        John Lin: ``Yes, I have noticed that. I think we need to keep an eye on him and make sure he's okay.''\\
        Mei Lin: ``Agreed. And speaking of neighbors, do you know Tamara Taylor who lives on the next block?''\\
        John Lin: ``I've heard of her, but I haven't had a conversation with her yet.''\\
        Mei Lin: ``Okay. And what about our elderly neighbors, the Moores? You've been acquainted with them for a few years, right?''\\
        John Lin: ``Yes, I really like Jennifer Moore's art. She's very talented.''\\
        Mei Lin: ``That's nice. Alright then, I'll see you later. Have a good day at work.''\\
        John Lin: ``Thanks. You too, Mei Lin.''\\
        \\
        Time: 07:53\\
        John Lin: ``Hey Tom, how's it going?''\\
        Tom Moreno: ``Good morning John, just checking the news before starting my day.''\\
        John Lin: ``Yeah, I hear you. Have you heard anything about who might be running for the local mayor election next month?''\\
        Tom Moreno: ``Actually, I was just reading an article about that. There are a few candidates being considered but I haven't made up my mind yet.''\\
        John Lin: ``Well, if you hear anything else, let me know. By the way, did you see the new medications and treatments I've been learning about in my online classes? There's some really exciting stuff out there.''\\
        Tom Moreno: ``No, I haven't had a chance to look into that yet. It sounds interesting though. By the way, have you noticed anything strange with your neighbor's rebellious son lately?''\\
        John Lin: ``Actually, I have. I'm planning on keeping an eye on him and making sure everything is okay. It's good to know that we are both keeping an eye out for the well being of our community. Oh, and thanks for the extra stock you gave me the other day. It really helped out.''\\
        Tom Moreno: ``No problem, John. I always try to help out where I can. By the way, have you seen any of Jennifer Moore's new artwork? She's really been stepping up her game lately.''\\
        John Lin: ``Yeah, I actually saw her new piece at the gallery the other day. It was really impressive.''\\
        \\      
        Time: 08:07\\
        John Lin: ``Hi there, Jennifer! I couldn't help but notice your beautiful artwork. I'm really looking forward to seeing your upcoming exhibition. Have you finalized the date yet?''\\
        Jennifer Moore: ``Thank you, John! I appreciate your kind words. And yes, the exhibition is scheduled for next month. I've been quite busy preparing for it.''\\
        John Lin: ``That's great to hear. I remember Tom Moreno mentioning how much he enjoyed your last exhibition. By the way, I heard you've been mentoring young artists. That's really inspiring. Can you tell me more about it?''\\
        Jennifer Moore: ``Yes, I love mentoring younger artists. It's such a joy to see them find their own creative voice. I have a few mentees right now who are working on some really interesting projects.''\\
        John Lin: ``That's wonderful. I'm sure they're lucky to have you as their mentor. On another note, have you been following the local politics?''\\
        Jennifer Moore: ``Actually, I'm not really interested in politics. I prefer to focus on my art and my relationships with the people I care about.''\\
        John Lin: ``I see. That makes sense. Well, it was really nice chatting with you, Jennifer. I'll be sure to mark the date of your exhibition on my calendar.''\\
        Jennifer Moore: ``Likewise, John. Thank you for stopping by. And please do come to my exhibition. I'd love to see you there.''\\
        \\
        Time: 09:16\\
        Giorgio Rossi: ``Excuse me, are you John Lin?''\\
        John Lin: ``Yes, that's me. How may I help you?''\\
        Giorgio Rossi: ``I noticed that you're very knowledgeable about medications and treatments. I've been curious about some of the latest developments in that field. Would you mind sharing some of your insights with me?''\\
        John Lin: ``Of course, I'm happy to share what I know. What specifically are you interested in?''\\
        Giorgio Rossi: ``Well, I'm particularly intrigued by how some medications are being developed based on mathematical patterns found in nature. Do you have any knowledge about that?''\\
        John Lin: ``Hmm, that's not really my area of expertise. But I do know that Jennifer Moore, the artist who's running for local mayor, is also a scientist who incorporates mathematical patterns into her work. Maybe she would have more information on that.''\\
        \bottomrule
    \caption{Example dialogue set for conversation evaluation.}\\ 
    \label{tab:dialoguesample}
\end{longtable}

\end{document}